\documentclass[conference]{IEEEtran}
\IEEEoverridecommandlockouts    

\usepackage{multirow}
\usepackage{amsmath,amssymb,bm}
\usepackage[T1]{fontenc}
\usepackage[utf8]{inputenc}
\usepackage{lmodern}
\usepackage{graphicx}
\usepackage{float}
\usepackage{hyperref}
\usepackage{microtype}
\usepackage{cite} 

\title{\LARGE \bf Decentralized Multi-View, Multi-Zone Tracking\\
for Scalable Camera Networks}

\author{
\IEEEauthorblockN{David Voihanski and Mor Sinai}
\IEEEauthorblockA{\textit{Juganu}\\
Or Yehuda, Israel \\
\{davidv, mors\}@juganu.com}
\and
\IEEEauthorblockN{Ben Zion Bobrovsky}
\IEEEauthorblockA{\textit{Tel Aviv University}\\
Israel \\
bobrov@tauex.tau.ac.il}
}

\date{}

\begin{document}
\maketitle

\begin{abstract}
We present a decentralized multi-view tracking system for scalable camera networks. Single-view detections are projected to bird's-eye-view (BEV) coordinates and fused by independent multi-view trackers, each managing a dynamic spatial zone. To enable high scalability across massive camera networks, we introduce a decentralized layer that serves as the primary driver for trajectory unification, utilizing a locking protocol and periodic track-merging operations which we term \emph{global collate}. Crucially, this process is supported by a small delayed-output buffering mechanism that retroactively applies history fix-ups; this brief, configurable delay window absorbs asynchronous coordination latency and has a major impact on stabilizing cross-zone identity consistency. We detail the design of dynamic zone polygons, consensus-based coordination, and this validation framework, which together ensure global ID consistency with minimal accuracy loss compared to centralized solutions.
\end{abstract}

\section{Introduction}

Multi-camera, multi-object tracking (MCMT) is critical for applications ranging from smart retail to city-scale security. Although recent work has advanced bird's-eye-view fusion~\cite{daryani2025camuvid, fleuret2008multicamera} and global ID consistency~\cite{tang2019cityflow, ristani2016performance}, many approaches rely on centralized compute nodes. These monolithic architectures not only struggle to scale to thousands of cameras on bandwidth-limited networks, but they also require substantial centralized processing infrastructure to handle the aggregated computational load.

We present a decentralized tracking system designed for such constraints, deployed on embedded devices using a three-layer architecture. First, a \emph{single-view} tracker runs per camera, outputting smoothed leg points. Second, a \emph{multi-view} tracker fuses these detections in a shared bird's-eye-view (BEV) space. Finally, a \emph{decentralized} layer partitions the site into dynamic BEV zones, each owned by an independent instance. These instances discover neighbors and coordinate via Raft-backed~\cite{ongaro2014search} decentralized locks to perform periodic \emph{global collate} operations that merge tracks across zone boundaries.

This paper focuses on the decentralized layer. By decentralizing coordination and keeping per-device processing requirements low, our system effectively scales to massive camera networks without central bottlenecks, maintaining a predictable computational footprint alongside highly competitive tracking accuracy.

\paragraph{Contributions} 
Our contribution is not a new single-camera tracker, BEV projection method, consensus protocol, or assignment solver. Instead, we study how to organize existing MCMT components into a decentralized tracking architecture that preserves cross-zone identity consistency under bounded per-instance computation. The methodological focus is the interaction between dynamic BEV zone ownership, overlap-based neighbor discovery, conservative cross-zone track collation, delayed output stabilization, and validation/reversion of provisional global merges. To realize this architecture, we make the following specific technical contributions:

\begin{itemize}
    \item A scalable decentralized design that partitions the BEV plane into dynamic zones handled by independent instances, eliminating the need for centralized servers;
    \item A mechanism for dynamically growing zone polygons and maintaining decentralized overlap map;
    \item A global collate pipeline that uses decentralized locking, hit-ratio gating, and force-threshold collision resolution to unify cross-zone trajectories;
    \item A novel validation and reversion layer that corrects erroneous global merges over time, integrated with a delayed-output buffer to ensure stable IDs.
\end{itemize}

\section{Related Work}

Single-camera tracking is well-studied, typically combining detection with Kalman filtering~\cite{bewley2016simple, wojke2017simple}. Multi-view approaches extend this to overlapping fields-of-view, often using early fusion in the bird's-eye (BEV) plane~\cite{daryani2025camuvid, fleuret2008multicamera} or location-aware matching at city scale~\cite{tang2019cityflow, ristani2016performance}. Although many systems rely on centralized servers, they face severe scalability limits due to bandwidth saturation and computational bottlenecks~\cite{kamal2013information, zhou2019edge}; thus, decentralized designs balancing local autonomy with coordination are increasingly relevant for large-scale deployments~\cite{kamal2013information}.

Our system is built on these foundations. Each camera runs a single-view tracker that outputs smoothed leg points in BEV coordinates. A local multi-view tracker fuses these points using an adaptive Kalman filter with fixed-lag smoothing. Unlike centralized approaches, we distribute tracking across independent instances that consume only local camera subsets. Coordination is handled by a lightweight layer using Raft-based~\cite{ongaro2014search} replication for decentralized dictionaries (mapping zone IDs to polygons) and a lock service for mutual exclusion. This ensures that the system eventually remains consistent without a single point of failure or a large central node.

\section{Problem Setting and System Assumptions}

\begin{figure}[t]
    \centering
    \includegraphics[width=0.8\linewidth]{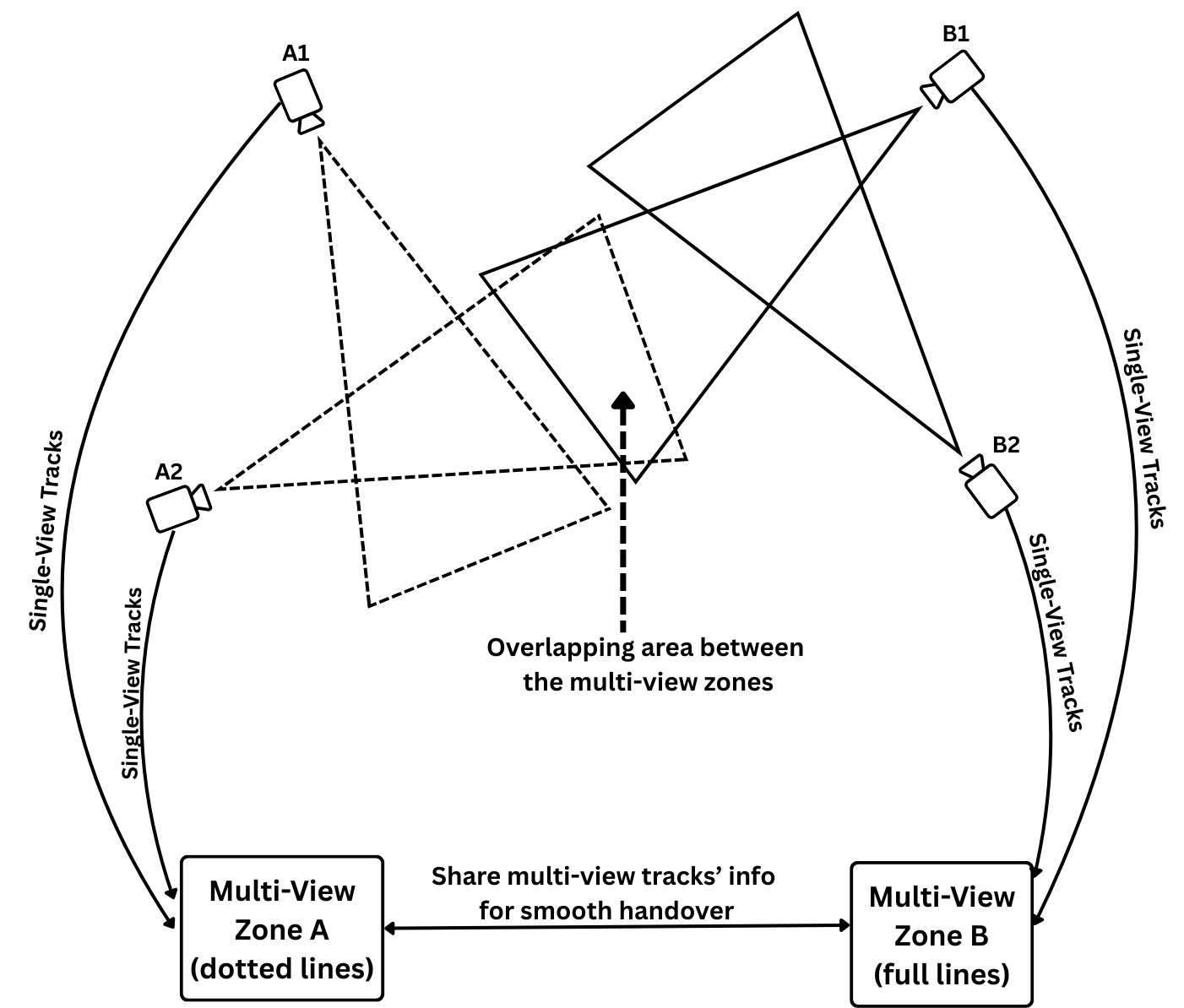}
    \caption{High-level architecture. Single-view trackers (outer arcs)
    run per camera and report smoothed leg points in BEV coordinates.
    One or more multi-view trackers operate on BEV detections within
    their zone polygons. A decentralized layer maintains overlapping BEV
    zones and coordinates global collate operations to maintain
    consistent track IDs across zones.}
    \label{fig:sv_multi_distributed}
\end{figure}

\begin{figure}[t]
    \centering
    \includegraphics[width=0.9\linewidth]{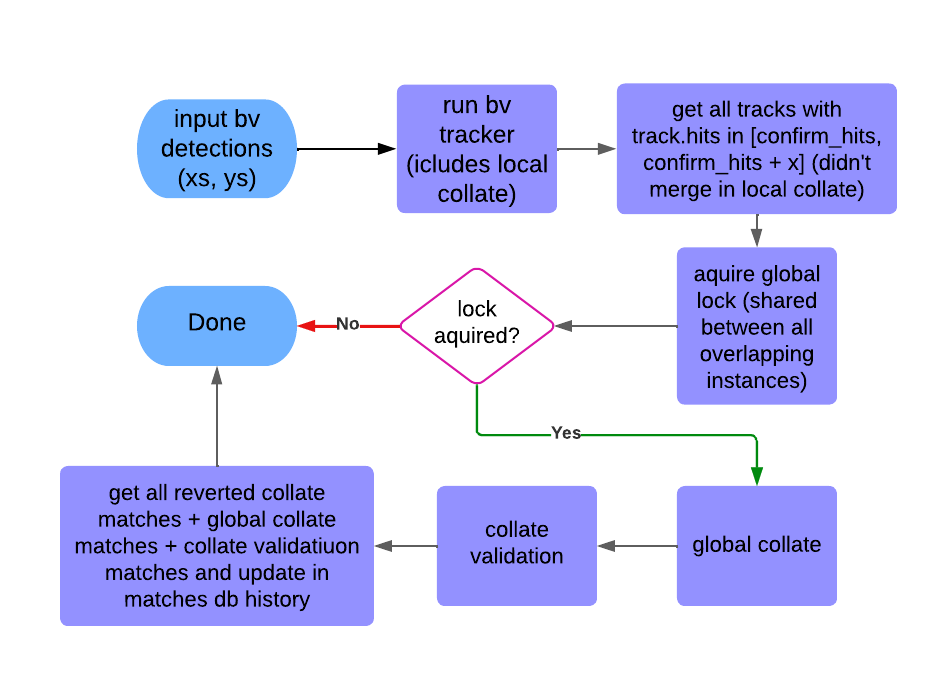}
    \caption{Decentralized multi-view tracker main loop. BEV detections are processed by the local BEV tracker. Tracks whose hit count lies in the global-collate window are collected, and the instance attempts to acquire the shared global lock for all overlapping zones. If the lock is obtained, it runs global collate and then collate validation; all accepted and reverted merges are written back into the match-history database before the iteration completes.}
    \label{fig:distributed_tracker_flow}
\end{figure}

Figure~\ref{fig:sv_multi_distributed} summarizes the hierarchy. We deploy a three-layer architecture: First, at the \textbf{single-camera} level, each camera runs a dedicated tracker that outputs a continuous stream of appearance embeddings and smoothed BEV leg coordinates. Second, local \textbf{multi-camera} instances fuse these detections in a shared BEV space. Finally, to scale, the \textbf{decentralized} layer introduces two key mechanisms: \textbf{dynamic BEV zone polygons}, which approximate coverage and enable neighbor discovery via intersection; and \textbf{global collate}, a locking protocol where neighbors periodically merge tracks crossing zone boundaries. 

The main loop of this decentralized process is detailed in Figure~\ref{fig:distributed_tracker_flow}. Crucially, we include a \textbf{validation and reversion} module that revisits these merge decisions using accumulated history, correcting errors before they persist in the output.

\section{Proposed Distributed Tracking Framework}

\subsection{Zone Polygon and Overlap Map}

\begin{figure}[t]
    \centering
    \includegraphics[width=0.9\linewidth]{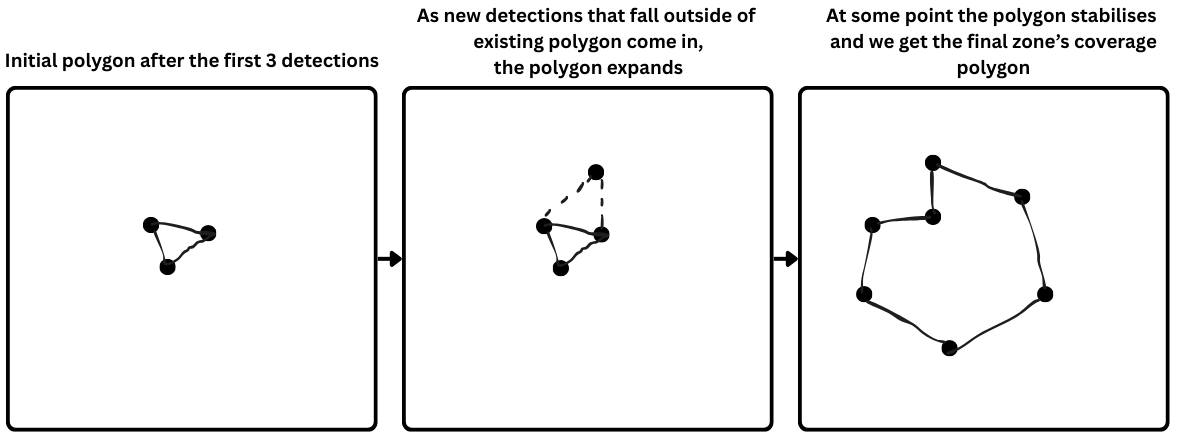}
    \caption{Dynamic zone polygon expansion. The multi-view zone starts as a small polygon around the first few detections (left). As new detections arrive outside the current polygon, the polygon expands to include them (middle). After sufficient coverage, the polygon stabilizes and represents the final spatial footprint of the zone in bird’s-eye coordinates (right).}
    \label{fig:zone_polygons_expansion}
\end{figure}

\begin{figure}[t]
    \centering
    \includegraphics[width=0.8\linewidth]{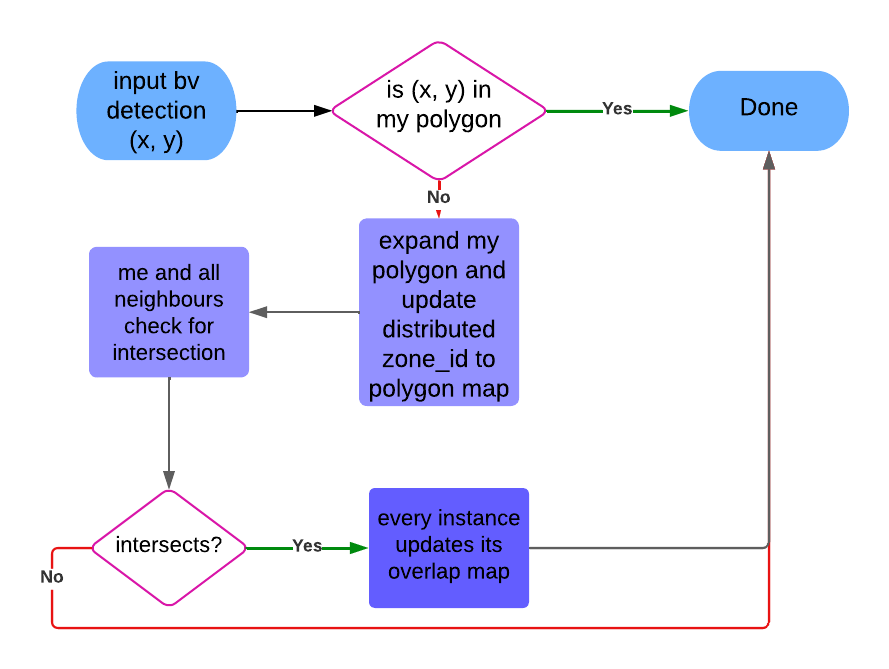}
    \caption{Zone polygon management in the decentralized tracker. For each incoming BEV detection $(x,y)$, the instance checks whether the point lies inside its current zone polygon. If not, it expands its polygon and updates the decentralized zone\_id~$\rightarrow$~polygon map. All instances then check for polygon intersections; when overlaps are detected, each instance updates its local overlap map so that neighboring zones are known for later global collate.}
    \label{fig:polygon_management}
\end{figure}

Each decentralized instance maintains a local BEV zone polygon $\mathcal{P}_i$ that adapts to the camera footprint, as shown in Figure~\ref{fig:zone_polygons_expansion}. After each tracking step, we collect BEV track centers; if a point falls outside the current polygon, we re-estimate $\mathcal{P}_i$ as the concave hull~\cite{moreira2007concave} of all accumulated points: $\mathcal{P} \gets \text{concave\_hull}(\{(x_k, y_k)\}, \text{ratio})$. This allows the zone to start small and gradually expand as the tracks explore the scene.

We synchronize these zones using a decentralized dictionary supported by Raft~\cite{ongaro2014search} $\text{polygon\_map}[i] = \mathcal{P}_i$. Instances periodically write updates if their local polygon changes and refresh their cache of the global map. Neighbor relationships are discovered by checking the geometric intersection with all other zones ($\mathcal{P}_i \cap \mathcal{P}_j \neq \emptyset$), as illustrated in Figure~\ref{fig:polygon_management}; overlapping instances are flagged as candidates for global collate.

\subsection{Decentralized Infrastructure}
Each instance initializes a local copy of the decentralized dictionary (for polygons) and a lock manager, which are collectively maintained and updated via distributed consensus directly among the participating tracking instances. A local RPC server provides services for global track matching and health monitoring. To minimize latency, instances maintain persistent RPC channels to all overlapping neighbors, enabling efficient asynchronous communication for collate operations without repeated connection setup.

\subsection{Global Collate Protocol and Locking}

\begin{figure}[t]
    \centering
    \begin{minipage}{0.48\linewidth}
        \centering
        \includegraphics[width=\linewidth]{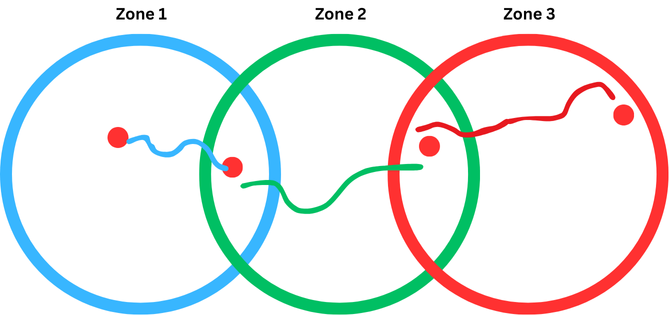}
        \centerline{(a) Without global collate}
    \end{minipage}\hfill
    \begin{minipage}{0.48\linewidth}
        \centering
        \includegraphics[width=\linewidth]{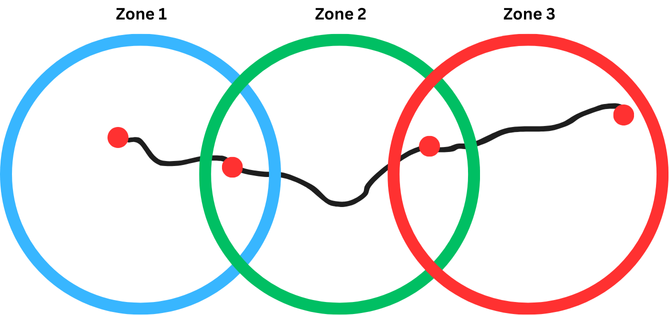}
        \centerline{(b) With global collate}
    \end{minipage}
    \caption{Impact of global collate on cross-zone tracking. (a) Without global collate, a person moving across overlapping zones generates fragmented local tracks with distinct IDs. (b) With global collate, local tracks are matched and merged, yielding a single consistent global ID for the entire trajectory.}
    \label{fig:global_collate_comparison}
\end{figure}

The cross-zone synchronization of tracking instances balances localized multi-view association with periodic global coordination. Figure~\ref{fig:global_collate_comparison} demonstrates the fundamental macroscopic impact of this decentralized mechanism: running independent trackers without cross-zone matching results in heavily fragmented trajectories at boundary intersections~(a), whereas executing our global collate protocol ensures complete trajectory unification across distinct zones~(b). 

\subsubsection{Execution Gating and Timers}
Global matching is attempted only when the shared-state engine reports a valid quorum and active neighbors are detected. To manage network load, operations are gated by two timers: a \emph{global collate interval} that limits the frequency of new merge attempts, and a \emph{collate validation interval} for re-verifying existing links. If either timer expires and newly confirmed tracks are present, the instance initiates the synchronization cycle by attempting to acquire the relevant communication locks.

\subsubsection{Global Lock}

To prevent race conditions, matching is protected by decentralized locks representing zone pairs (e.g., \texttt{1\_2}). The lock manager supports blocking acquisition with timeouts and robust release loops. When initiating global collate, an instance must acquire \emph{all} relevant locks for its overlaps; if any acquisition fails within the timeout, it releases all held locks and skips the cycle. This ensures that only one instance actively modifies the shared state of an overlapping region at any time.

\subsubsection{Collate Protocol and Matching}
Once locks are held, the instance sends serialized recently confirmed track histories (timestamps, Kalman states, embeddings) to neighbors via RPC. The receiving instance aligns these histories temporally and computes a pairwise association cost matrix between remote tracks and local candidates. This base matching cost is a weighted combination of a spatial kinematic metric and a visual appearance metric.

For the spatial evaluation, we compute the Mahalanobis distance $C_{\text{maha}}$ to account for state uncertainty and ensure kinematic feasibility. Let $x$ and $y$ be the position vectors of the local and remote tracks, and $S$ be the projection of the local track's Kalman filter covariance matrix into the measurement space:
\begin{equation}
    C_{\text{maha}}(x, y) = \sqrt{(x - y)^T S^{-1} (x - y)}
\end{equation}

For the appearance evaluation, we determine identity consistency by computing the Euclidean distance $C_{\text{emb}}$ between the high-dimensional visual Re-ID features $e_x$ and $e_y$ extracted by the single-view trackers:
\begin{equation}
    C_{\text{emb}}(e_x, e_y) = \|e_x - e_y\|_2
\end{equation}

The final pairwise association cost $C_{\text{total}}$ is computed as a weighted sum of the kinematic and appearance costs, parameterized by a balancing weight $\lambda \in [0, 1]$:
\begin{equation}
    C_{\text{total}} = \lambda C_{\text{maha}} + (1 - \lambda) C_{\text{emb}}
\end{equation}

To prevent impossible associations, candidate pairs that exceed a maximum Mahalanobis distance or a maximum feature distance are strictly gated by being assigned an infinite cost. The receiver then solves the linear assignment problem using the Hungarian algorithm~\cite{kuhn1955hungarian} to identify optimal track pairs from the fused cost matrix and returns them. The initiator applies the accepted matches by relabeling local track IDs and logging the merge timestamp for future validation.

\subsection{Validation and Reversion}

\begin{figure}[t]
    \centering
    \includegraphics[width=0.9\linewidth]{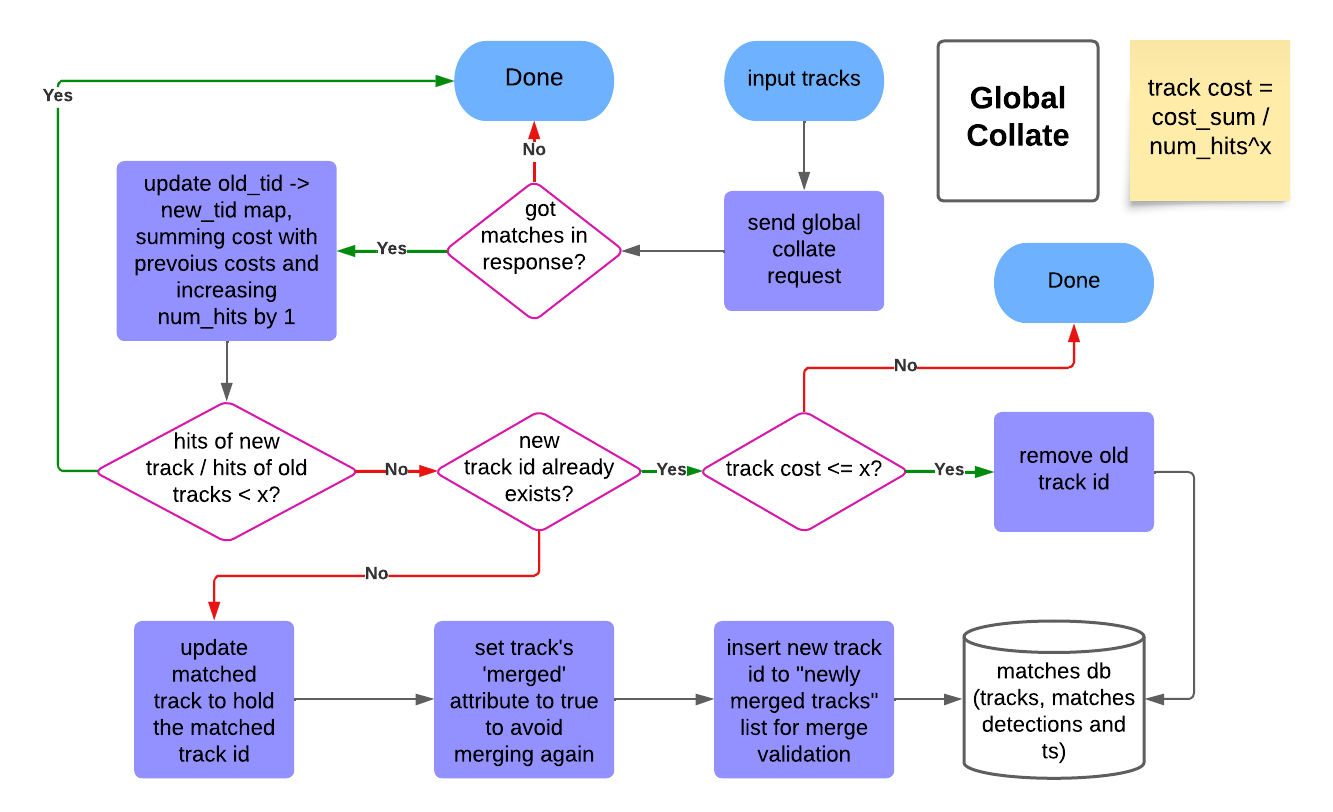}
    \caption{Global collate decision flow. Candidate matches must first pass a minimum hit-ratio check. If the incoming ID is unique locally, the merge is accepted immediately. If the ID already exists (conflict), the system forces the swap—removing the old ID—only if the match cost satisfies a strict force threshold.}
    \label{fig:global_collate_flow}
\end{figure}

\begin{figure}[t]
    \centering
    \includegraphics[width=0.8\linewidth]{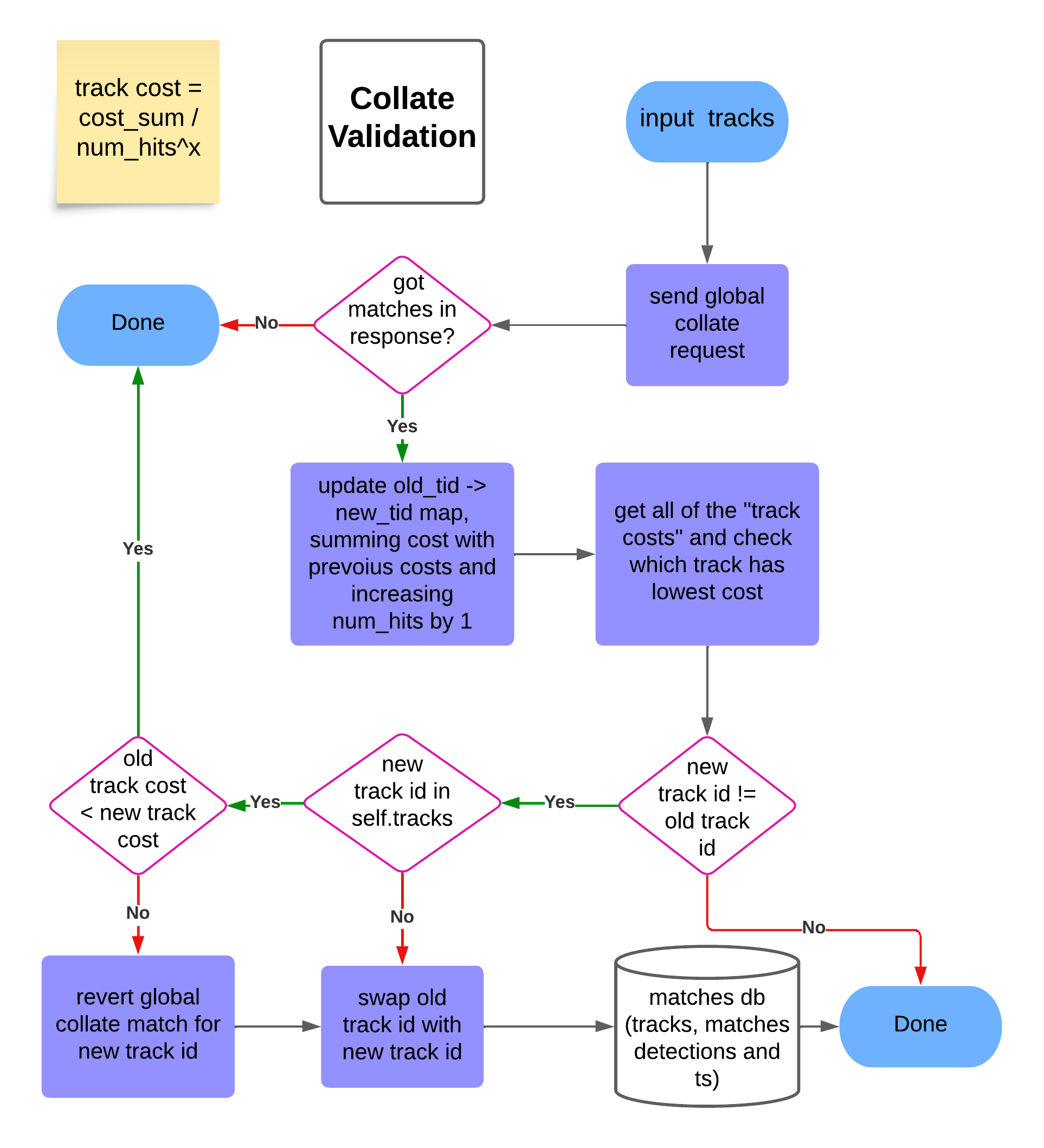}
    \caption{Collate validation logic. For a set of input tracks, the instance sends a global collate request and examines the returned matches. For each matched pair it accumulates hit counts and cost sums, builds an old\_tid~$\rightarrow$~new\_tid map, and then compares the validated cost of the old track vs.\ the candidate new track. Depending on this comparison, the system either reverts the global collate match or swaps the local track ID and updates the matches database.}
    \label{fig:collate_validation}
\end{figure}

For every proposed match between a local track \texttt{old} and
an incoming identity \texttt{new}, we accumulate association hits
and matching costs to evaluate the consistency of the proposed merge.

A \emph{hit} denotes a detection successfully associated with a
track during a tracking update. Let $h(t)$ denote the accumulated
number of hits for track $t$. Proposed merges must pass a
\textbf{Hits Ratio} gate:
\begin{equation}
    \frac{h(\texttt{new})}{h(\texttt{old})}
    \geq \tau_{\text{ratio}},
\end{equation}
where $\tau_{\text{ratio}}$ is the minimum required hit ratio.

If the incoming identity already exists locally, an ID collision
occurs. To decide whether to force the reassignment, we compute
the normalized consistency cost:
\begin{equation}
    J(a,b) =
    \frac{\mathrm{costs\_sum}_{a,b}}
    {(\mathrm{hits}_{a,b})^{m_{\mathrm{force}}}},
\end{equation}
where $\mathrm{costs\_sum}_{a,b}$ is the accumulated matching cost,
$\mathrm{hits}_{a,b}$ is the number of supporting associations,
and $m_{\mathrm{force}}$ controls the normalization by hit count.
Lower values indicate better matching consistency.

A forced swap $\texttt{old}\rightarrow\texttt{new}$ is permitted
only when $J(\texttt{old},\texttt{new})$ is below the configured
force threshold and is lower than
$J(\texttt{new},\texttt{new})$.

Merges are provisional. We track "recently merged" IDs and accumulate a hit-weighted score $S(c)$ for all candidates. If a candidate $c$ consistently outscores the current ID ($S(c) < S(\text{current})$), we trigger a \textbf{swap}. Collisions during this swap attempt to revert the conflicting track to its previous ID; if safe reversion fails, the conflicting track is removed.

\subsection{Output Delay Buffer and History Fix-up}
\label{sec:response_delay}
The decentralized tracker utilizes a temporal buffering and history-fix layer to manage the asynchronous time scales of global operations and absorb network coordination overhead.

\paragraph{Buffering Mechanism}
In each frame, the local tracking states and match hypotheses are queued alongside their corresponding timestamps into a fixed-length FIFO buffer. Rather than emitting track coordinates immediately, a background \emph{history-fix} routine continuously iterates over this buffered window to retroactively propagate consistency updates. This routine applies local ID re-mappings, modifies frames based on newly confirmed tracks, and executes global collate merges, forced swaps, or validation-driven reversions. Once the oldest buffer entry exceeds the configured decentralized delay, it is pushed to downstream systems as the official, stabilized output.

\paragraph{Configurability and Latency Absorption}
The system's output latency is fully configurable to balance tracking accuracy against real-time application constraints. While the response delay can be dialed down to zero seconds for ultra-low-latency tracking, operating without a buffer exposes the system to transient identity "flickering" at zone boundaries. This flickering occurs because provisional local IDs are briefly pushed to the live output before the asynchronous RPC coordination threads can return and retroactively unify the cross-zone tracks. 

Holding the stream back by a configurable window allows the system to accumulate temporal evidence, resolve spatial ambiguities, and hide global synchronization updates completely out-of-view. As detailed in our sensitivity evaluations, the tracking accuracy benefits of this buffer saturate at approximately \textbf{3.0 s}. Utilizing a primary evaluation operating point of \textbf{5.0 s} provides a highly conservative stabilization layer that ensures an entirely flicker-free identity stream.

\section{Experimental Setup}
\label{sec:experiments}

\subsection{Dataset Selection and Rationale}
Despite the prominence of existing benchmarks, they present specific limitations for evaluating the performance of a decentralized, scalable system. For instance, the \textbf{AI City Challenge}~\cite{tang2019cityflow} introduces domain shifts with non-target classes (e.g., industrial forklifts) and relies on a single averaged score, preventing us from isolating pedestrian tracking performance. In \textbf{MMPTrack}~\cite{han2023mmptrack}, restricted access to withheld test set labels precludes fully independent quantitative analysis of cross-zone ID consistency. Finally, while \textbf{WILDTRACK}~\cite{chavdarova2018wildtrack} provides high density, its brief 400-frame duration necessitates a train/test split (typically 360/40) that is insufficient for rigorously stress-testing decentralized locking protocols or accumulating drift over time. Consequently, we conducted our quantitative evaluation on large-scale \textbf{proprietary internal data}. This dataset mirrors real-world deployment conditions, allowing us to validate global ID stability and decentralized coordination far beyond the temporal and domain constraints of existing public benchmarks.

\subsection{Proprietary Datasets}
We utilize two proprietary multi-camera datasets collected in real-world retail environments. Due to privacy and business constraints, the raw videos and annotations cannot be publicly released. All datasets are recorded at \textbf{5 FPS} using fixed, \textbf{time-synchronized} cameras. 

The Ground Truth (GT) is labeled with \textbf{global identities across all cameras}, meaning a single individual observed in multiple views is assigned a unique, consistent GT track ID regardless of the camera or zone.

The characteristics of these datasets are summarized below:

\paragraph{Dataset A (Supermarket)} 
This dataset contains \textbf{11 cameras} and spans \textbf{2:20} minutes. Cameras are installed at an approximately \textbf{45$^\circ$} view angle and a height of roughly \textbf{3 m}. It includes \textbf{62} global ground-truth tracks and \textbf{16,854} labeled GT detections.

\paragraph{Dataset B (General store)} 
This dataset contains \textbf{8 cameras} and spans \textbf{2:50} minutes. Cameras are installed at an approximately \textbf{45$^\circ$} view angle and a height of roughly \textbf{4 m}. It includes \textbf{45} global ground-truth tracks and \textbf{14,943} labeled GT detections.

\section{Results and Analysis}

\subsection{Accuracy Evaluation}
We evaluate the accuracy of our decentralized system by comparing a \textbf{Single-Instance} baseline, where one node processes all cameras, against decentralized configurations where the deployment is partitioned into \textbf{Two-Instance} and \textbf{Three-Instance} setups. Unless otherwise specified, the reported full-system decentralized results use a \textbf{5.0 s} response delay.

Table~\ref{tab:results} summarizes the resulting HOTA~\cite{luiten2021hota}, IDF1~\cite{ristani2016performance}, and Identity Switches (IDSW)~\cite{bernardin2008evaluating} scores. As detailed later in the cross-zone attribution analysis accompanying Table~\ref{tab:id_attribution}, this performance drop under the 3-instance split is a consequence of restricted local view density rather than boundary coordination failures. On Dataset A, the two-instance run reduces HOTA by 1.4\% and IDF1 by 0.9\%, while IDSW remains unchanged at \textbf{4} in both the single- and two-instance runs. Splitting Dataset A into a three-instance configuration reveals an accuracy degradation roughly linear with the instance count, dropping HOTA by approximately 2.6 percentage points. Furthermore, the total IDSW doubles from 4 to 8 in the three-instance run. However, as detailed in the cross-zone analysis (Table~\ref{tab:id_attribution}), these additional identity errors are intra-zone rather than boundary failures. This amplified degradation is largely a consequence of over-partitioning a relatively small deployment: dividing Dataset A's 11 cameras across three instances leaves each node with only 3 to 4 cameras. This severely restricts the available field of view and multi-view redundancy within each local zone, making it significantly harder for the local tracker to resolve occlusions independently before tracks reach a boundary.

Dataset B is effectively flat in HOTA and slightly higher in IDF1, with IDSW increasing from 24 to 26. Together, these results show that the decentralized configuration preserves most of the centralized tracker’s identity quality, but that the accuracy trade-off increases as the site is partitioned into more zones.

\begin{table}[t]
    \centering
    \caption{Accuracy Comparison at 5.0 s Delay. Note that the 3-instance partition substantially reduces local multi-view redundancy (to only 3–4 cameras per instance), resulting primarily in localized intra-zone tracking dropouts rather than decentralized cross-zone matching failures (see also Table~\ref{tab:id_attribution}).}
    \label{tab:results}
    \setlength{\tabcolsep}{3pt}
    \footnotesize
    \begin{tabular}{l|c|ccc}
        \hline
        \textbf{Dataset} & \textbf{Configuration} & \textbf{HOTA} ($\uparrow$) & \textbf{IDF1} ($\uparrow$) & \textbf{IDSW} ($\downarrow$) \\
        \hline
        \hline
        \multirow{3}{*}{\textbf{Dataset A}} 
        & 1 Inst. (All)   & \textbf{0.911} & \textbf{0.962} & \textbf{4} \\
        & 2 Inst. (Split) & 0.897 & 0.953 & \textbf{4} \\
        & 3 Inst. (Split) & 0.885 & 0.942 & 8 \\
        \hline
        \multirow{2}{*}{\textbf{Dataset B}} 
        & 1 Inst. (All)   & \textbf{0.810} & 0.906 & \textbf{24} \\
        & 2 Inst. (Split) & 0.809 & \textbf{0.908} & 26 \\
        \hline
    \end{tabular}
\end{table}

\subsection{Component Ablation and Sensitivity Analysis}
To validate the contributions of the decentralized handoff machinery, we isolate the impact of global collate and delayed history fix-up. In the \textbf{No Global Collate} setup, zones operate independently, so tracks are broken and reassigned new IDs when they cross zone boundaries. The \textbf{Without Response Delay ($0$s)} setup enables global collate and validation but emits each frame immediately, before the delayed buffer can retroactively apply merge and reversion decisions. The \textbf{Full System} uses global collate, validation, and the 5.0 s delayed history fix-up used in Table~\ref{tab:results}.

\begin{table}[t]
    \centering
    \caption{Ablation Study on Decentralized Components (2 Instances)}
    \label{tab:ablation}
    \resizebox{\linewidth}{!}{%
    \begin{tabular}{l|l|ccc}
        \hline
        \textbf{Dataset} & \textbf{Configuration} & \textbf{HOTA} ($\uparrow$) & \textbf{IDF1} ($\uparrow$) & \textbf{IDSW} ($\downarrow$) \\
        \hline
        \hline
        \multirow{3}{*}{\textbf{Dataset A (2 Inst.)}} 
        & No Global Collate  & 0.635 & 0.706 & 48 \\
        & w/o Response Delay & 0.869 & 0.937 & 13 \\
        & \textbf{Full System} & \textbf{0.897} & \textbf{0.953} & \textbf{4} \\
        \hline
        \multirow{3}{*}{\textbf{Dataset B (2 Inst.)}} 
        & No Global Collate  & 0.735 & 0.855 & 64 \\
        & w/o Response Delay & 0.779 & 0.891 & 43 \\
        & \textbf{Full System} & \textbf{0.809} & \textbf{0.908} & \textbf{26} \\
        \hline
    \end{tabular}%
    }
\end{table}

As shown in Table~\ref{tab:ablation}, the largest accuracy gain comes from enabling global collate: IDSW drops from 48 to 13 on Dataset A and from 64 to 43 on Dataset B even when the output delay is zero. Delayed history fix-up then removes transient handoff errors that would otherwise leak into the output stream, further reducing IDSW from 13 to 4 on Dataset A and from 43 to 26 on Dataset B. This supports the core collate claim more directly than the aggregate single-vs.-decentralized comparison: global collate prevents most cross-zone fragmentation, and the delay buffer determines how many provisional merge decisions are stabilized before scoring.

\paragraph{Delay sensitivity}
Table~\ref{tab:delay_sensitivity} reports the two-instance delay sweep. The zero-delay row corresponds to immediate output with global collate enabled. Increasing the delay from 0 to 5 s improves HOTA by 2.8 percentage points on Dataset A and 3.0 percentage points on Dataset B, while reducing IDSW from 13 to 4 and from 43 to 26, respectively. The gains saturate at approximately 3 s on both datasets; the 5 s operating point is therefore conservative rather than necessary for all deployments.

\begin{table}[t]
    \centering
    \caption{Delay Sensitivity for Two-Instance Decentralized Runs}
    \label{tab:delay_sensitivity}
    \setlength{\tabcolsep}{4pt}
    \footnotesize
    \resizebox{\linewidth}{!}{%
    \begin{tabular}{l|c|ccc}
        \hline
        \textbf{Dataset} & \textbf{Delay (s)} & \textbf{HOTA} ($\uparrow$) & \textbf{IDF1} ($\uparrow$) & \textbf{IDSW} ($\downarrow$) \\
        \hline
        \hline
        \multirow{5}{*}{\textbf{Dataset A}} 
        & 0.0 & 0.869 & 0.937 & 13 \\
        & 1.0 & 0.887 & 0.947 & 10 \\
        & 2.0 & 0.896 & 0.951 & 9 \\
        & 3.0 & \textbf{0.897} & \textbf{0.953} & \textbf{4} \\
        & 5.0 & \textbf{0.897} & \textbf{0.953} & \textbf{4} \\
        \hline
        \multirow{5}{*}{\textbf{Dataset B}} 
        & 0.0 & 0.779 & 0.891 & 43 \\
        & 1.0 & 0.800 & 0.902 & 40 \\
        & 2.0 & 0.808 & 0.907 & 36 \\
        & 3.0 & \textbf{0.809} & \textbf{0.908} & \textbf{26} \\
        & 5.0 & \textbf{0.809} & \textbf{0.908} & \textbf{26} \\
        \hline
    \end{tabular}%
    }
\end{table}

\paragraph{Global collate accounting.}
Table~\ref{tab:collate_accounting} breaks down the outcomes of all global collate proposals at the 5.0 s operating point. To clarify the system's decision boundaries, every proposal is assigned to exactly one primary outcome. \textbf{Proposals:} The total number of cross-zone track merge attempts initiated by the system. \textbf{Accepted:} Clean merges where the proposed global ID was freely available in the local tracking instance, resulting in a successful unification. \textbf{Skipped:} Proposals discarded early because the remote match was too weak, specifically failing the minimum hits-ratio threshold. \textbf{Forced:} Merges where the proposed global ID already existed locally (a collision), but the association cost was confident enough to overwrite and replace the local track owner. \textbf{Collision Rej.:} Proposals where a collision occurred, but the merge cost either exceeded the force threshold or was worse than the track's self-assignment, causing the proposal to be safely discarded. \textbf{Validation No-op:} A second-pass validation of an earlier, previously accepted merge that confirmed the link and resulted in no assignment changes. \textbf{Reversions:} Cases where an earlier merge is undone during a subsequent validation pass. (Note: These are counted separately from initial proposals, as they represent a retroactive correction). \textbf{Failed:} Infrastructure or communication failures during the coordination cycle, specifically defined as cases where the instance either (1) cannot acquire the required global locks within the configured timeout window, or (2) encounters an RPC connection or transmission failure when communicating with a neighboring node. Notably, these failures were zero across all tested datasets.

\begin{table*}[t]
    \centering
    \caption{Global Collate Proposal Outcomes at 5.0 s Delay}
    \label{tab:collate_accounting}
    \setlength{\tabcolsep}{3pt}
    \footnotesize
    \resizebox{\textwidth}{!}{%
    \begin{tabular}{l|c|c|c|c|c|c|c|c}
        \hline
        \textbf{Dataset} & \textbf{Proposals} & \textbf{Accepted} & \textbf{Skipped} & \textbf{Forced} & \textbf{Collision Rej.} & \textbf{Validation No-op} & \textbf{Reversions} & \textbf{Failed} \\
        \hline
        \hline
        \textbf{Dataset A (2 Inst.)} & 81 & 36 (44\%) & 24 (30\%) & 0 & 1 (1\%) & 20 (25\%) & 0 & 0 \\
        \textbf{Dataset A (3 Inst.)} & 163 & 74 (45\%) & 68 (42\%) & 6 (4\%) & 12 (7\%) & 3 (2\%) & 5 & 0 \\
        \textbf{Dataset B (2 Inst.)} & 88 & 34 (39\%) & 44 (50\%) & 5 (6\%) & 5 (6\%) & 0 & 2 & 0 \\
        \hline
    \end{tabular}%
    }
\end{table*}

\subsection{Boundary Traffic and Cross-Zone Attribution}
    To thoroughly evaluate the reliability of our decentralized handoff layer, we separate our boundary analysis into two focused evaluations. First, Table~\ref{tab:boundary_profile} establishes the physical traffic and baseline stress profile across the zone boundaries, capturing the spatial behavior of the ground-truth targets. Second, Table~\ref{tab:id_attribution} isolates the exact attribution of identity switches to determine if our decentralized coordination introduces trajectory tracking errors. 

    As shown in Table~\ref{tab:boundary_profile}, the datasets present a highly active boundary environment; for instance, the three-instance split of Dataset A forces 34 separate ground-truth trajectories to navigate a total of 165 directional zone crossings, with crossing tracks spending an average of over 1,100 frames inside the overlapping boundary regions.

    \begin{table}[t]
        \centering
        \caption{Ground-Truth Boundary Traffic and Overlap Profile}
        \label{tab:boundary_profile}
        \resizebox{\linewidth}{!}{%
        \begin{tabular}{l|c|c|c|c}
            \hline
            \textbf{Dataset \& Config} & \textbf{GT Tracks} & \textbf{Crossing GT} & \textbf{Total Cross.} & \textbf{Mean Overlap Fr.} \\
            \hline
            \hline
            \textbf{Dataset A (2 Inst.)} & 62 & 28 & 137 & $\sim$844 \\
            \textbf{Dataset A (3 Inst.)} & 62 & 34 & 165 & $\sim$1155 \\
            \textbf{Dataset B (2 Inst.)} & 45 & 25 & 59  & $\sim$547 \\
            \hline
        \end{tabular}%
        }
    \end{table}

    \begin{table}[t]
        \centering
        \caption{Hierarchical Identity Switch Attribution Breakdown}
        \label{tab:id_attribution}
        \resizebox{\linewidth}{!}{%
        \begin{tabular}{l|c|c|cc}
            \hline
            \multirow{2}{*}{\textbf{Dataset \& Config}} & \textbf{System Total} & \textbf{Intra-Zone} & \multicolumn{2}{c}{\textbf{Cross-Zone IDSW}} \\
            & \textbf{IDSW} & \textbf{IDSW} & \textbf{Collate-Corr.} & \textbf{No-Match} \\
            \hline
            \hline
            \textbf{Dataset A (2 Inst.)} & 4  & 2  & \textbf{0} & 2 \\
            \textbf{Dataset A (3 Inst.)} & 8  & 6  & \textbf{0} & 2 \\
            \textbf{Dataset B (2 Inst.)} & 26 & 26 & \textbf{0} & 0 \\
            \hline
        \end{tabular}%
        }
    \end{table}

    The isolated identity metrics in Table~\ref{tab:id_attribution} evaluate the behavioral safety of our architecture by tracking how identity switches accumulate across different spatial regions. As shown by the horizontal breakdown, the total tracking errors (\emph{System Total IDSW}) are mathematically split into localized baseline failures (\emph{Intra-Zone IDSW}) and boundary handoff tracking drops (\emph{Cross-Zone IDSW}). Out of all evaluated configurations, the number of boundary switches actively caused by the matching loop (\emph{Collate-Corr. IDSW}) is \textbf{strictly zero}. These empirical findings demonstrate that our global locking protocol and validation framework introduced zero erroneous cross-zone identity splits or corrupted swaps across all evaluated sequences, confirming its structural safety under our test environments. 

    Crucially, when Dataset A is further partitioned from two to three instances—subjecting the coordination layer to maximum boundary traffic—there is strictly zero increase in identity switches correlated with global collate or validation events. The boundary errors that do persist are entirely unaccounted for by the matching loop (\emph{No-Match IDSW}), meaning the system occasionally fails to resolve a valid cross-zone association (false negatives) rather than actively generating destructive track modifications. On Dataset B, despite 59 boundary crossings, even these missed associations are completely eliminated, leaving only the baseline intra-zone errors. This precise attribution indicates that our localized coordination framework achieves high stability under evaluation, ensuring that increasing the multi-zone partitioning does not trigger trajectory corruption.

\subsection{Computational Scalability and Trade-offs}

\begin{figure}[t]
    \centering
    \includegraphics[width=0.95\linewidth]{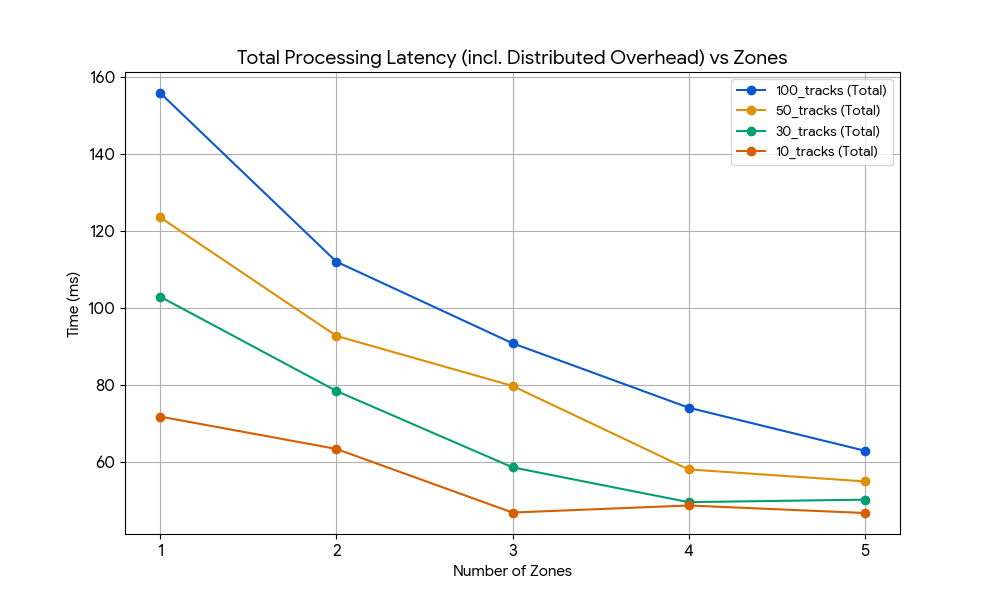}
    \caption{Total processing latency vs. number of zones ($N=40$ cameras). Data is generated with 100\% overlap (worst-case), yet latency improves as zones are added.}
    \label{fig:runtime_scalability}
\end{figure}

\begin{figure}[t]
    \centering
    \includegraphics[width=0.95\linewidth]{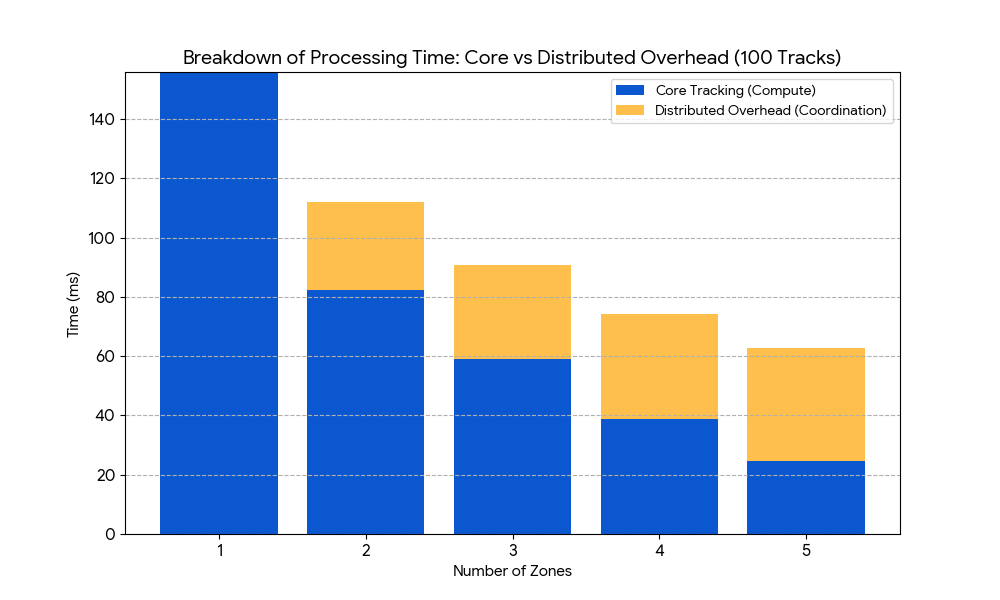}
    \caption{Latency breakdown (100 tracks, 40 cameras). As zones increase, core tracking time (blue) decreases faster than decentralized coordination (orange) increases.}
    \label{fig:overhead_breakdown}
\end{figure}

\paragraph{Computational Scalability} With $N_{\text{cam}}$ cameras and $N_{\text{zone}}$ zones, per-device load scales as $N_{\text{cam}}/N_{\text{zone}}$. We validated this using generated data (random features on BEV paths) as large-scale indoor datasets are unavailable. We configured a worst-case scenario with 40 cameras having 100\% field-of-view overlap, meaning every track is visible to every camera and triggers global coordination.

Figure~\ref{fig:runtime_scalability} shows that even under these worst-case conditions, dividing the site into more zones significantly reduces total latency. As shown in Figure~\ref{fig:overhead_breakdown}, while the decentralized coordination overhead (orange) increases with the number of zones, the reduction in the core tracking compute per zone (blue) is much steeper. This confirms that the system effectively distributes the load, eliminating central bottlenecks.

\paragraph{Scale-out interpretation}
We use the term scalable in the architectural sense: the computational input to each tracking instance is determined by its assigned zone, not by the total number of cameras in the deployment. If the site is partitioned so that each zone contains at most (C) cameras and overlaps with at most (K) neighboring zones, then local multi-view tracking and cross-zone collate remain bounded per instance as additional zones are added. In this regime, increasing the total camera count increases the number of instances rather than the size of a single centralized tracking problem. This scale-out model does require additional hardware as new zones are introduced, since each zone executes on its own tracking instance. However, the scalability results in figure~\ref{fig:runtime_scalability} and~\ref{fig:overhead_breakdown} show that partitioning the deployment significantly reduces the computational cost of each instance compared with processing the same cameras in a larger centralized zone. As zones are added, the per-instance workload decreases while existing instances maintain a bounded computational cost. This reduction in per-instance processing load is what enables the scale-out behavior described above: larger deployments are accommodated by adding zones and corresponding hardware resources rather than increasing the computational burden of existing tracking instances. Practical deployment limits are therefore determined primarily by available infrastructure, network capacity, synchronization reliability, and the density of zone overlaps rather than by a single centralized tracking bottleneck.

Looking forward, we hypothesize that under strict infrastructure and structural limitations, this decentralized architecture can theoretically achieve true unlimited scale. Specifically, if a deployment maintains a rigid cap on the number of overlapping zone boundaries per instance, and incorporates a proper allocation of localized ethernet access points for every $X$ embedded devices, network and locking contention can be kept entirely localized. While an empirical validation of this absolute scaling boundary is not proven in this paper, these structural conditions suggest a clear blueprint for scaling decentralized tracking across arbitrarily large camera networks without encountering local tracking overloads or global synchronization bottlenecks.

\paragraph{Accuracy Trade-offs} The decentralized design replaces a single global optimization problem with local tracking plus neighbor-to-neighbor cross-zone association. This makes per-instance computation depend primarily on local zone size rather than total deployment size, but it also introduces accuracy trade-offs that do not appear in a monolithic tracker. \textbf{Indirect Multi-hop Fusion:} Fusion across non-adjacent zones relies on iterative propagation rather than simultaneous global optimization. \textbf{Temporal Lag:} ID resolution is bound by the locking cadence, introducing a slight delay in unifying tracks across boundaries. \textbf{Boundary Effects:} Tracks that skim the edges of overlapping zones may occasionally miss merge opportunities.
We mitigate these effects by sizing overlap regions to provide sufficient temporal evidence for cross-zone association and by using conservative matching thresholds. This design reduces erroneous merges, but it can also leave some boundary tracks unmatched when the available evidence is insufficient.

\section{Limitations}
While our architecture demonstrates strong potential for scalability and identity preservation, its current scope has specific boundaries that require further study, which we categorize below into system and dataset constraints.

\paragraph{System Limitations}
First, while our architectural framework is designed to scale computationally, its empirical evaluation on physical deployments remains bounded. Our largest scale-out experiment ($N=40$ cameras) relies on generated trajectories to assess processing latency and computational throughput, leaving its end-to-end tracking accuracy at that scale unverified. Furthermore, the current real-data physical evaluation is restricted to a maximum of three decentralized instances. Stress-testing and tuning the underlying synchronization mechanics across massive, complex multi-zone layouts with a significantly higher count of physical instances remains an open engineering challenge for future work.

\paragraph{Data Constraints}
Second, our framework introduces strict constraints regarding both data availability and environmental layout. The annotated real-world datasets utilized in this work are restricted in camera count and duration (utilizing 8 to 11 cameras for under 3 minutes per sequence) because global multi-camera identity labeling is highly resource-intensive. More fundamentally, our tracking model imposes a strict structural requirement for spatiotemporal data continuity. The system relies entirely on continuous, intersecting zone boundaries to dynamically discover neighboring nodes and execute localized coordination. Consequently, the core algorithm cannot inherently bridge completely disjoint, fragmented data streams or spatially isolated tracking regions without incorporating an auxiliary wide-area re-identification layer.

\section{Conclusion}
We have presented a decentralized multi-view tracking system that partitions the bird's-eye plane into dynamic zones, each handled by an independent multi-view tracker instance, and coordinates cross-zone track unification via decentralized maps and locks. Compared to a single-instance multi-view tracker, our architecture sacrifices a small amount of tracking quality in exchange for the ability to scale across massive camera deployments while maintaining a bounded per-device load. By distributing the workload and eliminating the reliance on high-end centralized servers, our approach ensures a low and predictable computational footprint.

The empirical findings demonstrate the exceptional reliability of the global collate mechanism, which successfully achieves dozens of clean cross-zone merges across diverse deployment configurations to drastically minimize trajectory fragmentation. Furthermore, registering zero lock-acquisition or communication failures under contention confirms that our peer-to-peer locking and synchronization protocols are highly robust under active conditions. Finally, the validation and reversion modules function effectively as a conservative safety net—safely correcting minor edge-case anomalies while preserving tracking continuity across instances. Ultimately, the interplay between localized tracking, synchronized boundary collation, and buffered history fix-ups establishes a clear, practical blueprint for scaling multi-camera identity consistency across arbitrarily large networks without central processing bottlenecks.

\bibliographystyle{IEEEtran}
\bibliography{references}

\end{document}